\documentclass[runningheads]{llncs}
\usepackage[T1]{fontenc}
\usepackage{graphicx}

\begin{document}
\title{Computational Concept of the Psyche}
%
%
\author{Anton Kolonin\inst{1,2,3}\orcidID{0000-0003-4180-2870} \and
Vladimir Krykov\inst{4}\orcidID{0009-0008-4930-4618}}
\authorrunning{A. Kolonin and V. Krykov}
%
\institute{Novosibirsk State University, Pirogova 1, 630090, Novosibirsk, Russia\\
\email{akolonin@gmail.com}\\
\url{https://nsu.ru} \and
Aigents, Pravdy 6, Novosibirsk, 630090, Russia\\
\url{https://aigents.com} \and
SingularityNET Foundation, Baarerstrasse 141, Zug, 6300, Switzerland\\
\url{https://singularitynet.io} \and
Domodedovo Business Partner, Moscow, Russia}
\maketitle              
\begin{abstract}
This article presents an overview of approaches to modeling the human psyche in the context of constructing an artificial one. Based on this overview, a concept of cognitive architecture is proposed, in which the psyche is viewed as the operating system of a living or artificial subject, comprising a space of states, including the state of needs that determine the meaning of a subject's being in relation to stimuli from the external world, and intelligence as a decision-making system regarding actions in this world to satisfy these needs. Based on this concept, a computational formalization is proposed for creating artificial general intelligence systems for an agent through experiential learning in a state space that includes agent's needs, taking into account their biological or existential significance for the intelligent agent, along with agent's sensations and actions. Thus, the problem of constructing artificial general intelligence is formalized as a system for making optimal decisions in the space of specific agent needs under conditions of uncertainty, maximizing success in achieving goals, minimizing existential risks, and maximizing energy efficiency. A minimal experimental implementation of the model is presented.

\end{abstract}
\section{Introduction and Related Work}

Attempts to digitize the psyche or build a computable systemic model of it have been undertaken by many scientists, beginning with Norbert Wiener \cite{wiener1961cybernetics}. But it was Wiener who articulated the main challenge in finding a solution: the complexity of interdisciplinary knowledge and communication. As we will demonstrate below, creating a systemic model requires knowledge in psychology (psychoanalysis), systems analysis, and the economics of production markets (microeconomics). And building a software system that models the psyche requires knowledge of systems programming.

In our view, the creation of an artificial psyche as a control system for the functioning of an artificial agent (AA) is a solution to the problem of creating an artificial general intelligence (AGI) equal to or superior to human intelligence. Human intelligence is a system that controls its activities, solving the fundamental problems of survival and reproduction. To carry out such activities, a system, whether human or AGI, must make optimal decisions at every moment in time and throughout its entire life cycle. According to the general definition of intelligence given in the earlier works \cite{goertzel2021generaltheorygeneralintelligence,wang2006}, intelligence is the ability to achieve complex goals in complex conditions with limited resources. At the same time, if the fulfillment of the resource constraint condition can be considered one of the sub-goals in the internal hierarchy of a complex composite global goal, then the presence of intelligence in a system can formally be reduced to the ability to implement multi-parameter optimization when solving a target control problem in a specific operating environment. In this case, the level of intelligence will be determined by the quantity and complexity of the parametric space structure of a given operating environment and the system's goals. In other words, the task of intelligence is to find optimal solutions at various levels of interaction with the surrounding world, regardless of whether the system is a living animal, a human, or an inanimate AA.

According to Kahneman \cite{kahneman2011thinking}, the human activity control system consists of two parts: fast "system 1" (reflexive or intuitive) and slow "system 2" (conscious thinking). Consciousness belongs to the slow part and exerts virtually no control over life processes. Consciousness makes some decisions and perceives only some external and minimal internal information about the state of the system.

Decision-making in a living system, as well as formation of its experience by means of learning process, is accomplished through emotions \cite{simonov2021emotional,dubynin2024brain}. Emotions are a mental reaction to the satisfaction of a need, the achievement of a goal, or failure to achieve it. Emotion shifts the system to a new state, compensating for a deviation from equilibrium. Positive emotion arises in the system when a need or goal is satisfied. Negative emotions arise when an unsatisfied need demands satisfaction. This serves as an informational signal to the system to increase the priority of the need. Pain signals perform the same function at the physiological level.

To create a model of intelligence that includes the agent's response to internal motivating stimuli and the perception of reactions from the external world, in the context of existing expectations and the agent's own influence on the environment, it is proposed to use Anokhin's theory of functional systems (TFS) \cite{Sudakov2015} and the principle of dynamic equilibrium \cite{bertalanffy1968general}. Based on the hierarchical TFS model, a cognitive architecture can be constructed that describes the behavior of both an autonomous intelligent agent \cite{VITYAEV2018623} and society as a whole \cite{KOLONIN2016475,kolonin2026}.

\section{Proposed Concept and Model}

\subsection{Anthropocentric Model of the Psyche}

\begin{figure}[t]
\centering
\includegraphics[width=0.85\columnwidth]{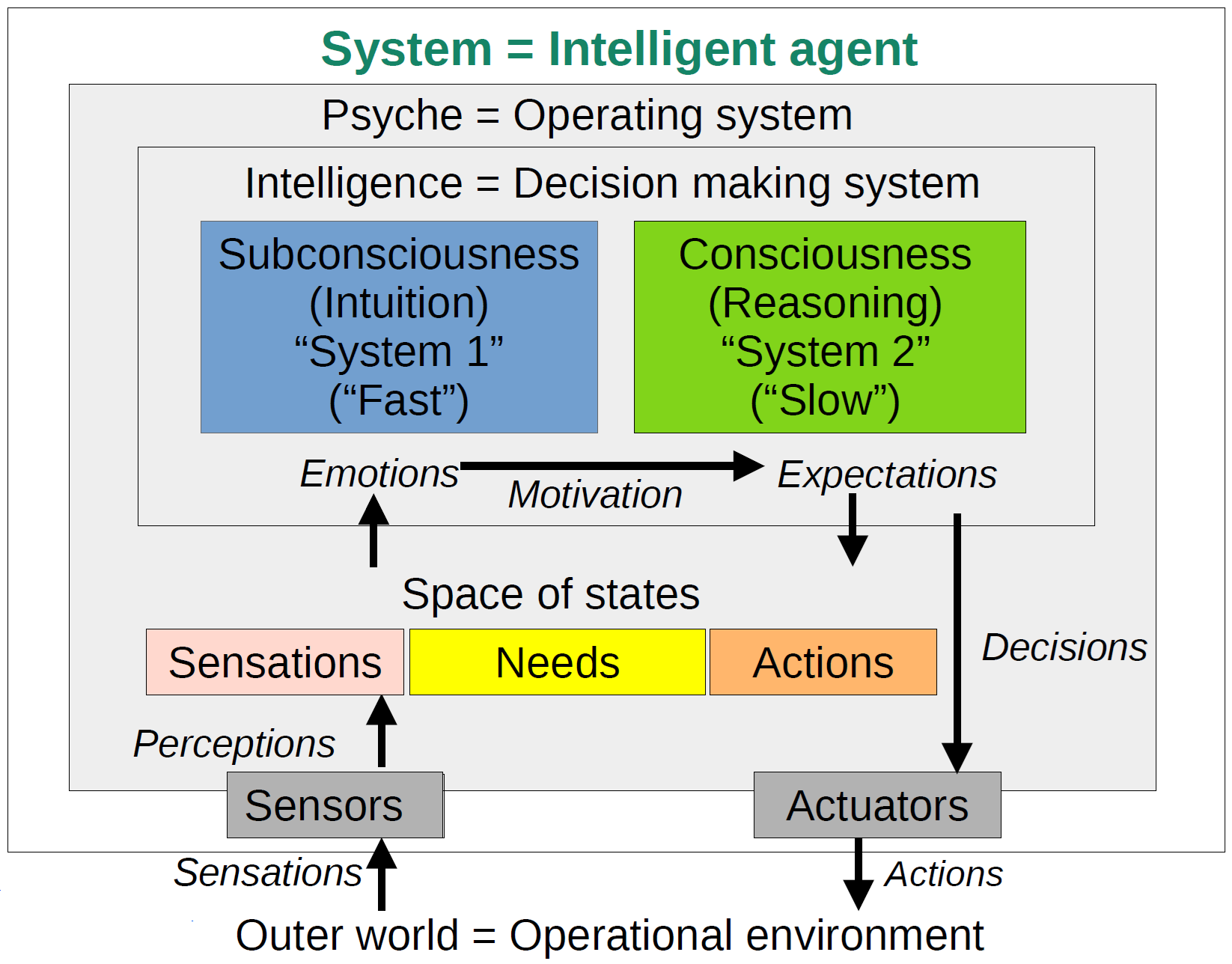}
\caption{Conceptual cognitive architecture of artificial psyche.}
\label{fig1_psyche}
\end{figure}

In a broader sense, intelligence is a system of analysis and decision-making based on knowledge and experience, ensuring the effective functioning of an individual, society, or a high-level intelligent system (including artificial intelligence agent), or a set of such systems. Moreover, intelligence is not an abstract set of laws of the physical world or facts derived from it, but rather the ability of a system to control its activity based on meaningful experience. Today, intelligence exists only in living systems, but we are exploring the possibility of creating it in artificial systems, proposing the following anthropocentric concept of the mind for the implementation of a computational model, as shown in Figure~\ref{fig1_psyche}.

The psyche is an operating system for managing human life processes, or a hypothetical artificial agent. Intelligence is its decision-making subsystem, comprising subconscious and conscious components that operate complementarily and competitively, as demonstrated in the work of Cisek2007. The subconscious makes reflexive decisions (as in the case of human intuition or the use of modern systems based on deep neural networks), while the conscious mind makes strategic, conscious decisions within a certain planning horizon, taking into account the availability of resources.

The process of making decisions to achieve current goals and satisfy urgent physiological and psychological needs can be described as motivation to perform certain actions under the influence of emotions. Motivation for action is determined by both external factors and the internal state of the system. The internal state is determined by the priority of physiological and psychological needs, as well as the level of available energy resources and external physical resources. The influence of the external environment, as well as data on the state of subsystems (organs of a living organism), are perceived by the system's sensors as sensations and transmitted as perceptions to change its internal state. Decisions made during the motivation process are transmitted for execution to the system's actuators to influence the environment or internal subsystems (organs of a living organism).

At any given moment, the system selects the most efficient solution to satisfy current needs if their priority is high. If absence of the current needs, it finds an activity that will effectively satisfy future needs that will inevitably arise. An intelligent system may know that the body will need food in a few hours, but the exact timing of hunger will depend on energy expenditure on activities in the near future. Therefore, the intelligent system strives both to provide the necessary resources that will soon be needed and, if possible, to reduce their consumption.

Computational and, especially, economic efficiency issues have been considered as important elements of AGI computing architectures before \cite{Wang1996,Kolonin2015,kolonin2026}. However, existing approaches in AGI, both those based on maximizing probabilistic predictions \cite{Friston2010} and on various types of probabilistic logic \cite{goertzel2008probabilistic,VITYAEV2013159}, clearly do not take efficiency considerations into account. At the same time, David and LeCun \cite{Dawid_2024} also recently pointed out the need to introduce the concept of energy efficiency as fundamental for constructing AI architectures. It should be noted that some existing models of probabilistic logic allow for implicit consideration of energy efficiency factors as additional variables or predicates \cite{goertzel2008probabilistic,VITYAEV2013159} while others provide account for them explicitly \cite{Wang1996,kolonin2026}.

First difference between the proposed concept and existing purely probabilistic models \cite{goertzel2008probabilistic,VITYAEV2013159} and models where resource consumption is a factor limiting the logical inference \cite{Wang1996,Kolonin2015,kolonin2026} is that the decision is assumed to be oriented towards the final result in the perspective of risk management, both existential (a threat to the existence of the system or the society in which it is located) and energy or economic, in light of prospect theory \cite{tversky1982judgment}.

Moreover, in the proposed model, any needs of the agent itself form a vector of different needs with potentially different levels of expression, which create a vector space of the agent's needs. These needs are placed in the context of the agent's sensations, which represent another vector, determined by the expressions of these sensations at any given moment and described by the vector space of sensations. Finally, the decision-making process may involve performing one or more actions simultaneously, so the combination of all possible actions performed by the agent forms the action space. The combination of these three spaces creates what we can call the agent's psyche space of states.

According to Anokhin \cite{Sudakov2015}, real intelligence is focused on the final result. In this regard, we believe that decision-making can be implemented using a computational method of economic cost analysis, taking into account, among other things, the probabilities of existential threats and the potential costs of preventing them. Thus, a solution can be selected based on the criteria of greatest utility, maximum safety, and minimum cost (the efficiency coefficient for obtaining survival energy) for the system based on multi-objective analysis. Anokhin \cite{Sudakov2015} came close to this, but the models of the Marxist planned Soviet economy he used could not reflect the competition of tasks in multi-objective systems with the need for risk management. By applying the laws of the market economy, the intelligence model becomes computable.

Ludwig von Bertalanffy's general systems theory \cite{bertalanffy1968general} assumes that a living system is in constant dynamic equilibrium. External influences or internal changes deviate the system from equilibrium. Any deviation requires a return to equilibrium. This requires action, an expenditure of system energy. And this energy must be replenished. We believe solutions to this problem can be found in multi-objective economic risk management models \cite{tversky1982judgment}.

To ensure the survival of a species, society, or individual, it carries out purposeful activity. This activity requires energy, which comes from food. Providing food is one of the most important tasks all living organisms perform, along with avoiding threats to their existence.

Providing food also requires performing certain actions. Thus, we obtain a global, closed cycle of life activity, based on survival energy. Consequently, "survival energy" should be used to evaluate goal-directed activity. Not in a direct physical sense, but as a "universal currency" (similar to the idea of an "energy dollar" or a "token" in digital crypto-economics) for comparison, as a multifaceted conventional unit that provides a quantitative assessment of the physiological and psychological processes of a computational model of a living organism or an artificial psyche as an operating system for an AGI agent.

Survival energy is the biological (in the case of a living organism) or electrical (in the case of an artificial system) energy required by an organism or system to perform purposeful external activities and carry out internal processes.

Bekhterev wrote about the energy of the body and psyche \cite{bekhterev1932general}. He described the various types of energy in the body and the principles of their interaction. At the beginning of the 20th century, Freud, in his book Beyond the Pleasure Principle \cite{freud1920beyond}, wrote that the psyche operates according to economic principles.

The internal state of the system is described in a space of needs, called the "matrix of needs" according to \cite{kryukov2023selflearning}, which is the main element of the model of a living intelligent system (e.g., a human) or a computational cognitive architecture of an AGI system (AA), describing the motivational segment of the agent in its space of states.

Within the proposed concept, system learning occurs through the acquisition of experience, which enables decision-making to meet the system's needs within a planning horizon. In the case of humans, human knowledge consists of a vast amount of elementary experience gained through one's own actions or from texts (the significant experiences of others). This knowledge, defined in the space of needs, is structured into a \cite{e12051145} coverage. Self-learning is about gaining and understanding experience and choosing solutions that are most effective in terms of both costs and benefits.

Adler \cite{adler1927understanding} discovered the basis for the formation of experience in the psyche during the process of personality development. Tendencies toward increasing the significance (psychological value) of goals or needs can be understood from his original idea of the "inferiority complex." And his concept of psychological value allowed for the comparison and calculation of psychological and physiological needs.

The notion of the experiential learning in artificial intelligence was suggested more that 20 years ago by Goertzel \cite{goertzel2004experiential} and later re-stated in \cite{kolonin2022} as a way to have an autonomous agent learning any knowledge from the outer world by its own experience and feedback received from the environment or inferred by the agent itself. It could be supervised learning with feedback obtained by implicit labels or directions, semi-supervised learning with feedback inferred from rewards and punishments provided by environment or instructor, or self-supervised learning where the feedback is inferred by an agent itself based on its expectations grounded on its internal model of the surrounding world.

\subsection{Space of Needs in Space of States}

Human intelligence develops every second from the moment of birth. Therefore, a model of intelligence should be built starting from the simplest state. An agent's intelligence operates not with real objects, including its own body, but with internal images of the surrounding world and the body within it as sensations, with the agent's own actions and needs — that is, a reflection of the world and the agent itself in its psyche. Such internal images, dynamically changing over time, can be described as a space of states: in the most primitive case, one-dimensional space, and in the case of real living organisms or when implemented in AGI systems with a complex hierarchy of interdependent sensations, actions and needs, a tensor space. The segment of space corresponding to the agent's needs and goals is called the space of needs, also known as the "needs matrix" in \cite{kryukov2023selflearning}.

To solve the fundamental problem — the survival of an individual, society, or species — the system's goal becomes the satisfaction of current needs in accordance with the chosen priority. The dynamics of functional needs can be described by the mathematics of market economics, including risk management through the probabilistic analysis of the outcomes of alternative scenarios \cite{tversky1982judgment}. The interdependence between physiological and psychological needs is calculated through the psychological value of the intellectual being as potential (future) energy for survival, or, more precisely, the satisfaction of basic needs/motivations, of which survival is only one.

The flow of needs is continuous. According to Maslow \cite{maslow1971farther}, the satisfaction of one need leads to the emergence of a new one; needs compete and are interdependent. Anokhin \cite{Sudakov2015} also pointed to the competition and interdependence of needs. A goal, in turn, differs from a need in that it is finite. A goal is complete when it is achieved, while a need requires satisfaction on ongoing basis.

The priority values of current needs determine the current state of the system's psyche. Current values of satisfaction, dissatisfaction, and demand (priority) of a need characterize the current point (coordinate) in the functional space of goal-directed human activity in the Markov coverage perspective \cite{e12051145}.

Higher-order needs can be expressed through a combination of basic needs \cite{maslow1971farther}. Therefore, any system goal can be described as a combination of basic needs and lower-order needs. The space (or "matrix" \cite{kryukov2023selflearning}) of needs is the fundamental element of the system, upon which all calculations and comparisons are based. Needs reflect the current state of the system. Essentially, the space of needs is a set of cells arranged according to the system's state levels (as in an operating system supervisor program), where each cell contains the current priority value of the need. The space of needs can be considered as a universal mathematical tool for describing the activity of a human model when describing their behavior or an artificial intelligence system when implementing it.

When modeling the human psyche, the space of needs, reflecting the significance of any object for humans in nature, consists of the possibilities for satisfying these needs at different levels. In this case, the space of needs can be decomposed into the following levels: a) Individual (basic); b) Family and household; c) Socioeconomic; d) National and civilizational. Each level has its own priority in decision-making, and priorities may differ for different people.

It appears possible to create a mathematical model that allows us to understand an agent's dynamic worldview as a space of its states, including sensations and actions, with its motivational point in the space of its needs. This makes it possible to find the optimal solution for the activity of a model of a living intelligent organism or an instance of an artificial intelligence system at any point in such space. It is possible to construct a decision-making algorithm based on selecting the most effective solution to the current problem, taking into account risks and resource constraints within a certain planning horizon.

\subsection{Mathematical Model}

\begin{figure}[ht]
\centering
\includegraphics[width=0.85\columnwidth]{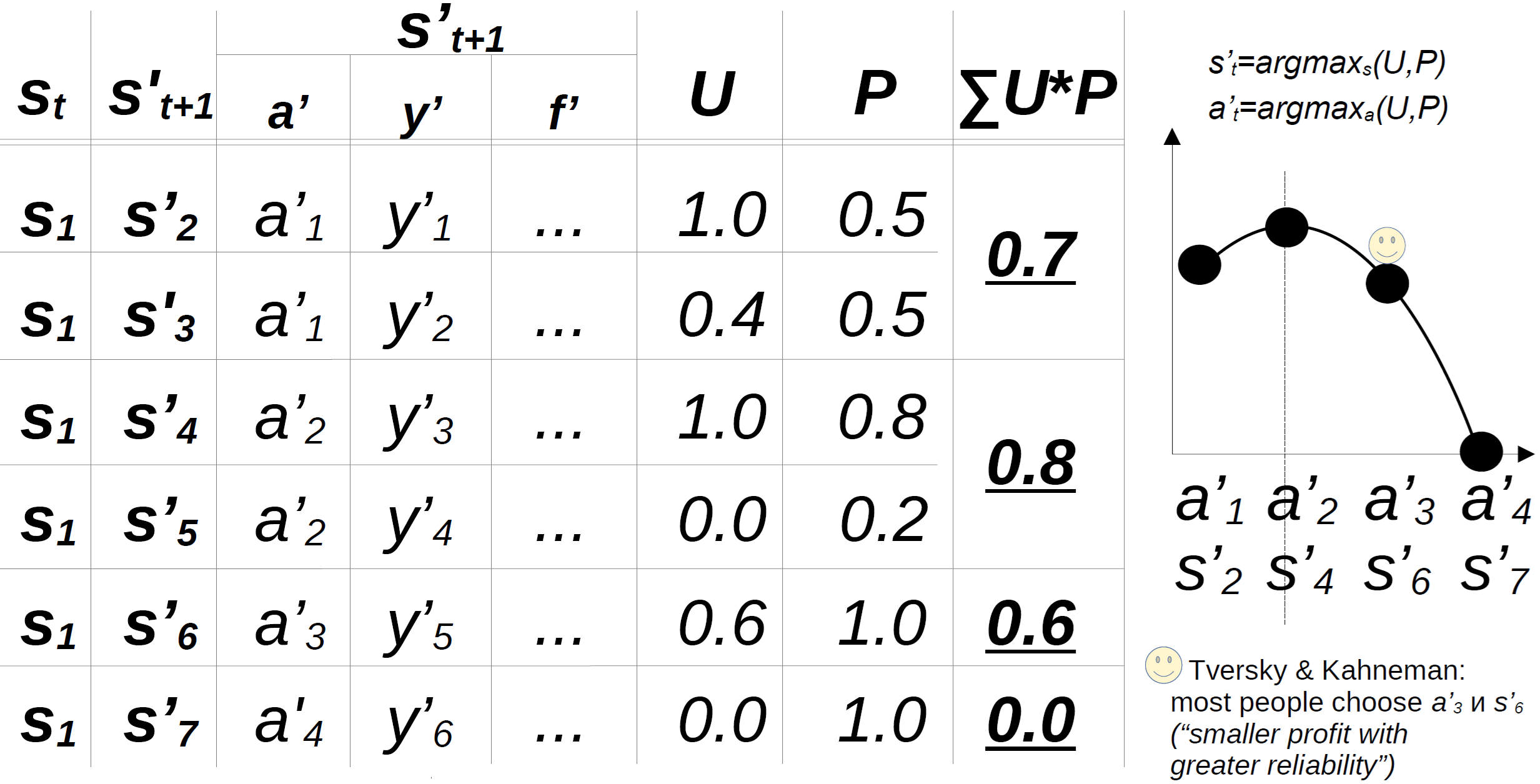}
\caption{
Illustration of the decision making process in state $s_t$ at time $t$, described by vector $s_1$, where 6 transitions to 6 states in the future time $t+1$ are possible, represented by vectors $s'_2$, $s'_3$, $s'_4$, $s'_5$, $s'_6$, $s'_7$, based on 4 possible actions $a'_1$, $a'_2$, $a'_3$, $a'_4$. In the case of actions $a'_1$ and $a'_2$, a non-deterministic transition to new states with different probabilities may occur. Action $a'_1$ may lead to $s'_2$ or $s'_3$, while $a'_2$ may cause $s'_4$ or $s'_5$. Actions $a'_3$ and $a'_4$ lead to a deterministic outcome. Each state vector includes a corresponding action vector $a'$, a feeling vector $f'$, and a need satisfaction vector $y'$, associated with the overall utility $U(s) = U(y)$ of transitioning to this state and the probability of this transition $P(s)$. Each action $a'$ can be associated with a statistically estimated "prospected utility" such as $\sum U * P$. Although statistical and economic evaluation based on "prospect theory" \cite{tversky1982judgment} suggests that action $a'_2$ is optimal (either maximum profit with moderate probability, or no profit with low probability), the same research shows that people usually intuitively choose actions like $a'_3$, where a smaller profit is guaranteed with higher probability.}
\label{fig2_utility_probability}
\end{figure}

If we consider the functioning of the human psyche throughout the life cycle, as well as the artificial psyche of a hypothetical intelligent agent, as learning based on experience of interaction with the environment, according to \cite{kolonin2022}, then in the context of the development of artificial intelligence systems, the most common formulation is reinforcement learning \cite{vanOtterlo2012}. It is based on the Markov model of the decision-making process \cite{puterman2005markov} and assumes than reinforcement received by an agent in response to an action taken against the environment is used to determine the so-called "utility" of the particular decision or action that preceded the reinforcement, in light of the "expected utility" theory \cite{von2004theory}. However, in modern economics, the latter theory has given way to the "prospect theory" \cite{tversky1982judgment}, according to which the utility of an action is assessed by both the positive and negative consequences of its implementation, as well as the various probabilities associated with these consequences, while the subjective utility of positive and negative consequences can be different, and these differences can be different for different subjects, as illustrated in Fig~\ref{fig2_utility_probability}.

In practice, when learning from experience, it is necessary to evaluate each decision or action not only in terms of individual reinforcement, but also in terms of a combination of possible explicit reinforcements, obtained upon achieving different goals (getting food or pleasure), or implicit ones, such as avoiding threats (electric shock or dousing with water), acquiring useful life experience, or reaching a certain social status, which may be necessary for achieving goals and avoiding threats in the distant future, but cannot be "reinforced" in the near term. Thus, an expansion of the scalar assessment of utility is required, replacing it with a vector one, as, for example, in \cite{kolonin2022}, utility is calculated in a two-dimensional space of positive and negative reinforcement — as the utility of obtaining gains and avoiding losses in accordance with "prospect theory" \cite{tversky1982judgment}. The need for explicit support for exploratory activity can consist in introducing informational utility for discovering new patterns and increasing the efficiency of energy consumption — due to energy saving and gaining utility. In complex systems, such as humans or imaginable highly intelligent agents, the space of needs can be structured into a heterarchy of interconnected needs and described not by a vector but by a matrix, where the rows of the matrix correspond to needs at different levels, from basic to higher \cite{maslow1971farther} or, more generally, by a tensor. Thus, the principle of intelligence as a mean to achieve complex goals, according to Goertzel \cite{goertzel2021generaltheorygeneralintelligence} and Wang \cite{wang2006}, can be formalized by defining a complex tensor utility function corresponded to a complex structured goal.

The concept of a complex goal, in the context of Fig.~\ref{fig1_psyche}, can be formalized as a "motivational vector" $z$ \cite{petrenko2012goal}. Based on the space of states that we have defined, with a space of needs being part of it, the motivational vector can be defined as the scalar product $z = x \cdot y$ of two vectors: the long-term vital importance (priority) of the need $x$ and the short-term dissatisfaction (actualization) of this need $y$. 

The long-term priority vector $x$ can be viewed as a constant genetic code of the agent, predetermined by its genetic makeup (in the case of humans) or inherently built-in (in the case of artificial intelligence agents). It can also be viewed as a semi-permanent cultural code of the agent, acquired through its development and long-term learning ("personality formation"). Partially, the latter part of it may be corresponding to agent's belief being subject of change due to social environment pressure \cite{kolonin2026}. It determines, for example, the ratio of propensities to take risks for the sake of acquisitions (positive reinforcements) and avoid losses (negative reinforcements) in light of prospect theory \cite{kahneman2011thinking}, the desire to ensure energy efficiency \cite{Sudakov2015,goertzel2008probabilistic,VITYAEV2013159,bekhterev1932general,freud1920beyond}, and the need for cognition, satisfied as an increase in the predictability of the surrounding reality, according to \cite{Friston2010}. That is, long-term priority $x$ can be thought of as an agent's "personality profile".

In turn, the actualization vector $y$ is determined on the basis of data obtained in the form of emotions \cite{simonov2021emotional,dubynin2024brain} as dissatisfaction of the corresponding needs at a specific moment $t$ of decision-making.

\begin{figure}[ht]
\centering
\includegraphics[width=0.99\columnwidth]{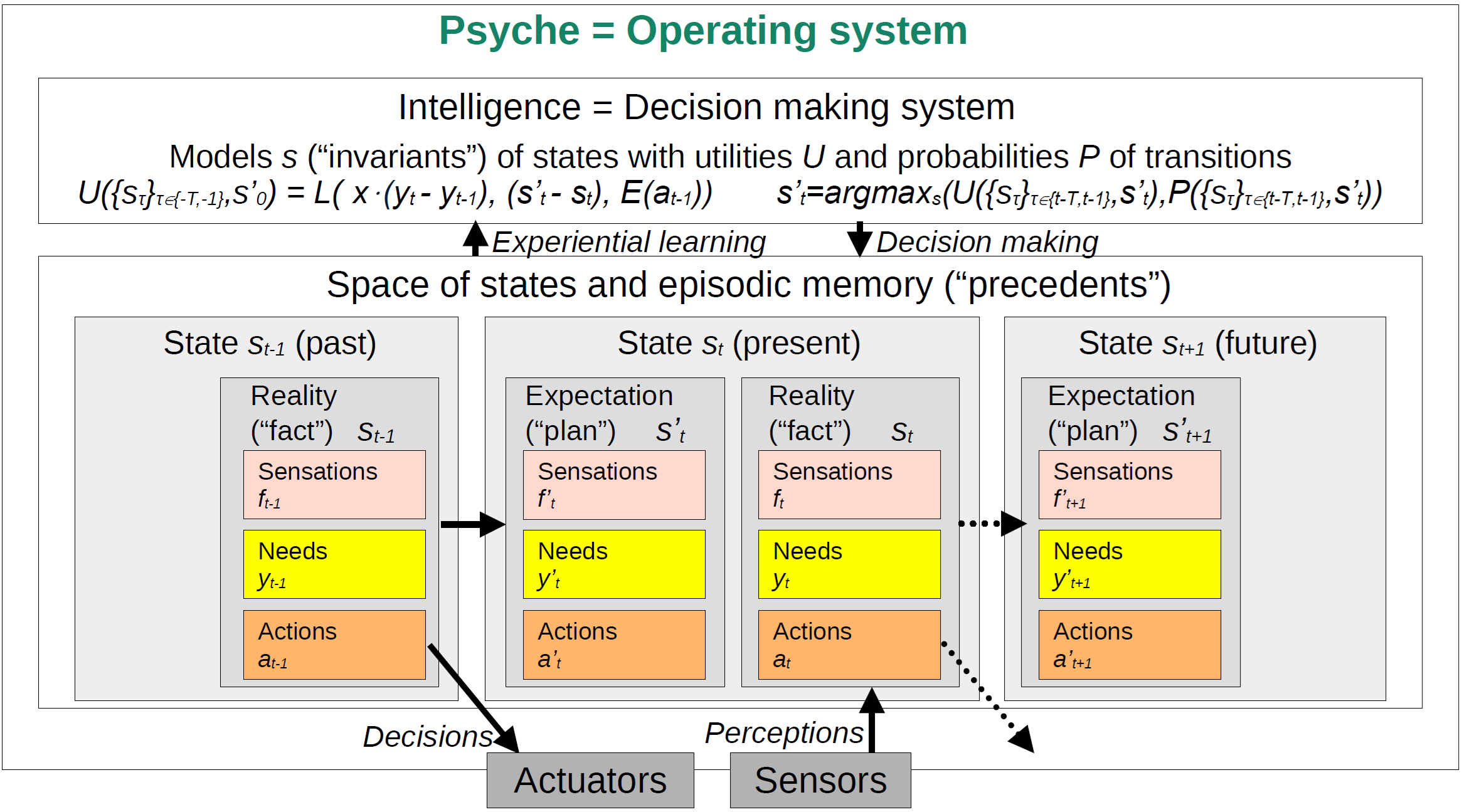}
\caption{
Illustration of the decision making process in the state of spaces based on state transition graphs stored in the episodic memory of an agent, where experienced states $s$ are agent's memories kept as precedents linked by transitions, associated with per-transition utility and evidence counts while predicted states $s'$ are expectations that an agent experiences at any given moment while making a decision or calmly observing an environment.}
\label{fig3_states}
\end{figure}

Thus, within the framework of the Markov model of the decision-making process \cite{puterman2005markov}, the expected value or utility of the implementation of a particular vector of actions and inactions $a$ in the current state $s$, for a moment in time $t$, as well as the reinforcement received during training \cite{kolonin2022,vanOtterlo2012} at the next moment in time $t+1$, can be determined by the following vector and scalar functions, as shown in Fig.~\ref{fig3_states}.

$s_t$ is a vector of state variables with values determined at a time $t$, according to Fig.~\ref{fig1_psyche}, by means of the agent's sensors, including: sensors of need actualization $y$, for example, those perceiving positive and negative reinforcements, the acquisition or loss of an energy resource, as well as confirmation of the expectation of the results of a prediction or action (acceptors of the result of an action \cite{Sudakov2015}); sensors of recording the performance of an action $a$; sensors of sensations or feelings $f$, not explicitly associated with needs or activity. $s'_{t+1}$ is a vector of values corresponding to the agent's expectations at time $t_1$, projected on the basis of its decision-making process.

$a_t$ is a vector of actions performed by the agent ($a \subseteq s$), with the perception of these actions is recorded by the corresponding state variables. Actions scheduled to be performed at the next time are defined as a committed action vector $a’_{t+1}$. In case if no any action is performed ($a$) or planned ($a'$), it corresponds to a zero vector.

$y_t$ is a vector of needs actualization, expressed through the dedicated state variables, where the needs actualization is part of the state ($y \subseteq s$), with the expected needs actualization in the planned future is denoted as $y’_{t+1}$. 

$f_t$ is a vector of sensations not directly related to the actualization or satisfaction of needs or the execution of actions ($f \subseteq s$), with the sensations anticipated in the future based is defined as $f'_{t+1}$. 

$x$ is a vector of needs prioritization (agent's personality profile), assumed to be constant for the agent in the short term, but can be changed in the long term (for instance, corresponding to gradual belief change over time according to \cite{kolonin2026}.

$z_t = x \cdot y_t$ is a vector of motivation to satisfy needs, taking into account their prioritization and actualization (“motivational vector” \cite{petrenko2012goal}).

$U(s,s’)$ is a scalar function (matrix) of the utility of experiencing the state $s’$ (including the actions $a’$ expected to be performed in this case or without performing any actions at all) for the initial state $s$ in the context of prioritizing the needs $x$ of the agent, determined in the course if learning based on implicit reinforcements $r_{t+1}$ upon transitions from the state st to $s_{t+1}$, corresponds to the Q-function in the case of using the Q-learning method \cite{li2023reinforcement} (in the latter case, the utility function degenerates into $U(s,a’)$). 

$C(s,s’)$ is a scalar function (matrix) of the evidence count of experiencing the state $s’$ past state $s$, and it can be used to compute respective probability distribution matrix for state transitions $P(s,s’)$.

$Q(s,s'')$ and $M(s,s'')$ are the matrices of mutual exclusion ($Q$) or mutual dependency ($M$) of the above variables $s$ which can be provided as an innate or hard-coded constraints to guide or restrict the decision making process based on utility and probability. $Q$ indicates if some of them cannot be physically satisfied ($y$), realized ($a$) or experienced ($f$) by the agent simultaneously. $M$ specifies which of them must be satisfied, accomplished or experienced simultaneously or one is necessary for the other.

In a non-Markov decision making model, the learned utility $U$, evidence count $C$ and accumulated probability $P$ can be assessed based on transitions between sequences of states $\{s_\tau\}$ and subsequent states $s’_0$ in time window of size $T$, like $U(\{s_\tau\},s’_0)$, where $\tau\in\{-T,-1\}$.

$r_{t+1} = x \cdot (y_t — y_{t+1})$ is a scalar reinforcement function (corresponding to emotional response in animals and human) resulting from action or inaction $a_{t+1}$ during the transition from state $s_t$ to state $s_{t+1}$, including the “special” needs: 1) predictability as a reverse function of the difference $s’_{t+1}-s_{t+1}$ (“expectation minus reality”) between expectations $s’_{t+1}$ formed at time $t$ and the actual state $s_{t+1}$ at time $t+1$; 2) energy efficiency $E(a)$.

Moreover, a system-wide utility learning function $L$, in which the explicit reinforcement $x \cdot (y_t - y_{t+1})$ is separated from the implicit reinforcements based on predictability and energy efficiency, can be rewritten as follows.

\[
U(\{s_\tau\},s’_0) = L( x \cdot (y_t  - y_{t+1}), s’_t  - s_t, E(a_t)), \tau\in\{-T,-1\}
\]

It is worth noting that the $L$ function given above, which has only the first argument, corresponds to the classical reinforcement learning problem statement, while the same function, which has only the second argument, corresponds to the classical problem statement for unsupervised learning of classical transformer models \cite{DBLP:journals/corr/abs-1810-04805}.

Based on the above, the decision making function that selects the expected and desired state $s’t$ given a sequence of preceding $T$  states $\{s_\tau\}\tau, \tau\in\{-T,-1\}$ can be based on maximizing the “prospected utility” based on the learned utility and probability distributions on the state transition graphs, as follows.

\[
s’t = argmax_s(U(\{s_\tau\}\tau,s’_t),P(\{s_\tau\}\tau,s’_t)), \tau\in\{t-T,t-1\}
\]

It is worth noting that in the case of the classical formulation of the reinforcement learning problem, the decision-making function is reduced to $s’t = argmax_s(U(\{s_\tau\}\tau,s’_t))$, whereas in the example in Fig.~\ref{fig2_utility_probability} it is assumed $s’t=argmax_s(U(\{s_\tau\}\tau,s’_t) \cdot P(\{s_\tau\}\tau,s’_t))$.

\subsection{Architecture}

We propose that the duality of the two systems described by Kahneman \cite{kahneman2011thinking} can be implemented in the so-called hybrid neuro-symbolic architecture \cite{kolonin2022}, which allows for intelligent activity to be carried out in one of two different model representations with the possibility of knowledge transfer between the two: 1) associative networks with latent knowledge, such as deep artificial neural networks and transformers, or 2) interpretable knowledge graphs or hypergraphs with probabilistic weights of statements of symbolic logic \cite{Wang1996,VITYAEV2013159,goertzel2008probabilistic}. The first can be associated with Kahneman's "System 1", and the second can be considered as "System 2".

Domingos's recent theoretical work, \cite{domingos2025tensorlogiclanguageai}, demonstrated that this is possible to have a knowledge or a model represented by any symbolic logic system, as well as by any modern neural network architecture, expressed in the so-called "tensor logic" language. Accordingly, any world model or world knowledge can be reinterpreted either as a heterarchical hypergraph with labeled or unlabeled nodes and edges, or as a distributed sparse representation in some hidden tensor space of some topology. And we propose that, given the presence of a formalism that unites "both worlds," the transformation and transition of the two models in an agent's mind is possible ib both directions.

Interestingly, the same knowledge structure in a tensor space or hypergraph can provide different options for improving inference efficiency and enabling life-long learning. On the one hand, if we store knowledge in the network as numbers on edges corresponding to learned probabilities or trained parameters, then reducing these numbers and pruning the network to 8, 4, and 2 bits leads to Boolean 1-bit parameters corresponding to the old, well-performing logic network. On the other hand, if we associate each point in the hypergraph element's tensor space with a compound or complex truth value, according to Wang \cite{Wang1996} and Goertzel \cite{goertzel2008probabilistic}, or with a complete evidence log or evidence counts per each observation, we can eliminate the possibility of catastrophic forgetting when perceiving new evidence due to the dis-balance between under-fitting and over-fitting within the model.

Such a conceptual design can be implemented in the technical architecture of an artificial psyche as the operating system of an artificial agent. The psyche, as the agent's internal world, consists of a space of dynamically changing states, constructed from the transient values of state variables corresponding to the agent's perception of the external world, the satisfaction of its current needs or goals, and confirmations of actions directed at the world, aimed at satisfying needs and achieving goals.

A decision-making system, whether neuro-associative, symbolic, or hybrid neuro-symbolic, analyzes the history of past states, uses a learned model of the world to make predictions based on various possible actions, finds appropriate future states that satisfy needs and achieve goals to the greatest extent, and selects the appropriate actions to perform.

The learning function, as described in the previous section, takes into account the relevant action outcomes in a historical contexts and updates the utility function values corresponding to transitions from a past sequence of states to future sequences, accumulating evidence counts of the observations for each transition to update estimations of the transition probabilities probabilities. The decision or inference function, in turn, at each new transition fork maximizes the total utility and the probability of choosing the most useful new state with the highest probability.

\begin{figure}[ht]
\centering
\includegraphics[width=0.8\columnwidth]{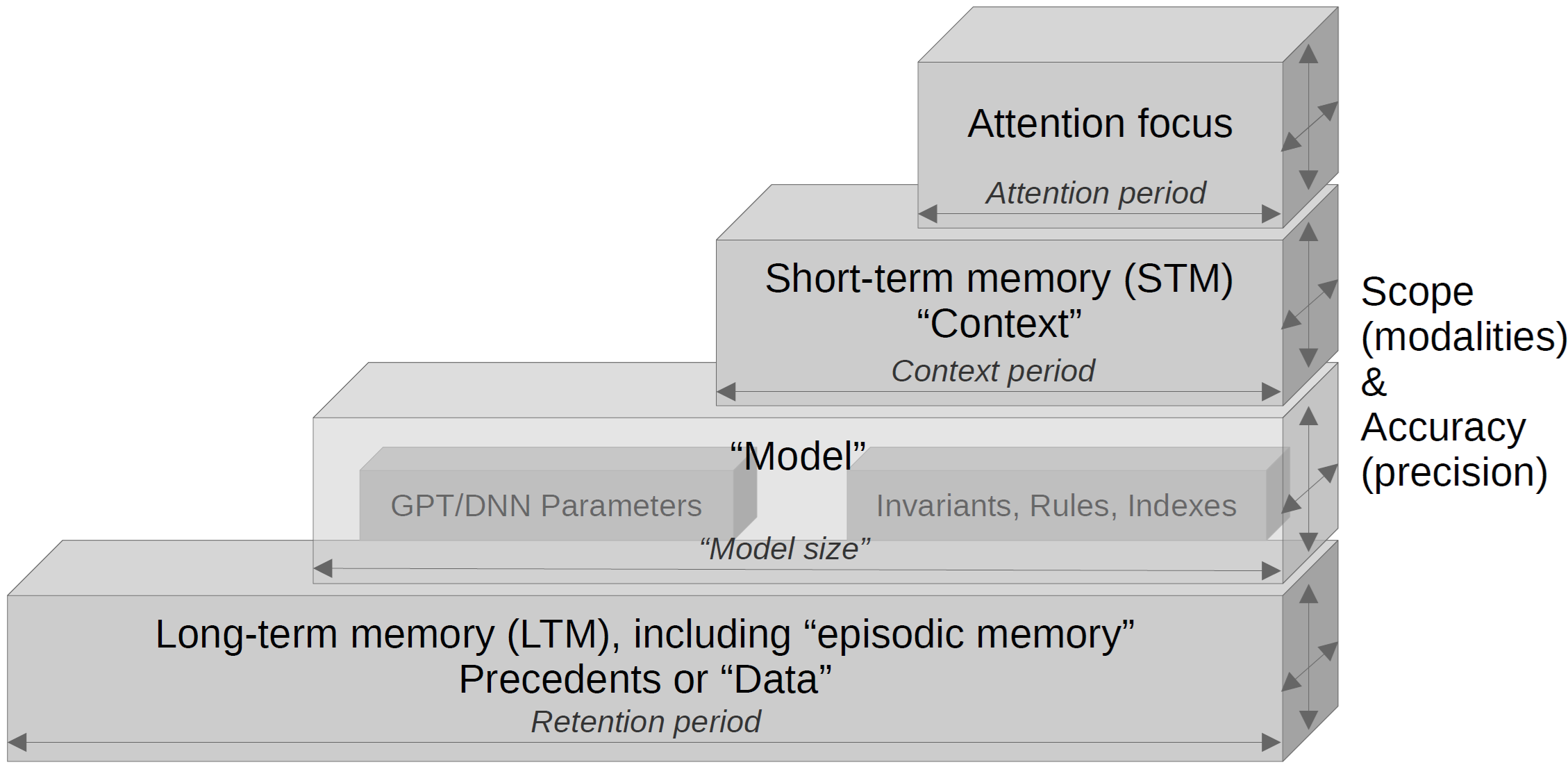}
\caption{Four-layer memory architecture (from bottom to top): 
a) long-term episodic memory keeping logs of agent's interactions with the environment; b) "model" of the agent-environment interactions that is either neural-associative (left) or symbolic (right); c) attention focus with current observations; d) short-term memory with current operational context. Memory data in each of the four layers can be limited in three dimensions: 1) size, capacity, or time period, limited to storing only recent events on certain retention horizon; 2) scope of evidence perception or number of input and action modalities; 3) precision or accuracy of the data (e.g., in the form of floating-point numbers, integers, or Boolean values).}
\label{fig4_architecture}
\end{figure}

The memory architecture of such a system can be four-layer and include the following elements, as shown in Fig~\ref{fig4_architecture}. First, episodic long-term memory, containing a complete experience log, is necessary to ensure a full accounting of evidence and prevent catastrophic forgetting. However, it can be subject to garbage collection for low-trust and outdated experiences. The second layer, the "model" memory, can be represented either as neural-associative network architectures such as transformer-based, or as symbolic models in the form of graphs, rules, and predicates of invariant patterns based on base precedents in long-term memory. This layer can be viewed either as an approximation of the base precedents or as an index for accessing them in form of invariants, as presented in \cite{Vityaev2022}. Next, the operational context used to form and guide decision-making and the acquisition of new knowledge can be represented by short-term memory. Finally, the current situation is stored in attention focus memory. One can imagine how this architecture might compare to some imaginable information retrieval-augmented generation architecture based on a large language model, where these layers are represented as follows: 1) a vector, graph, or relational database storing dynamic data and interaction logs used to periodically retrain the model; 2) the "model" itself; 3) a context window; 4) a prompt with a query as a focus of attention.

\section{Empirical Evaluation}

As an experiment, a minimal implementation of training an agent to play ping-pong against the opposite wall("Single-player Ping-Pong") was carried out, extending the experimental setup described and implemented in \cite{kolonin2022}; the open source code of the solution is published at \url{https://github.com/aigents/aigents-java/tree/master/src/main/java/net/webstructor/agi}.

\begin{figure}[ht!]
\centering
\includegraphics[width=0.8\columnwidth]{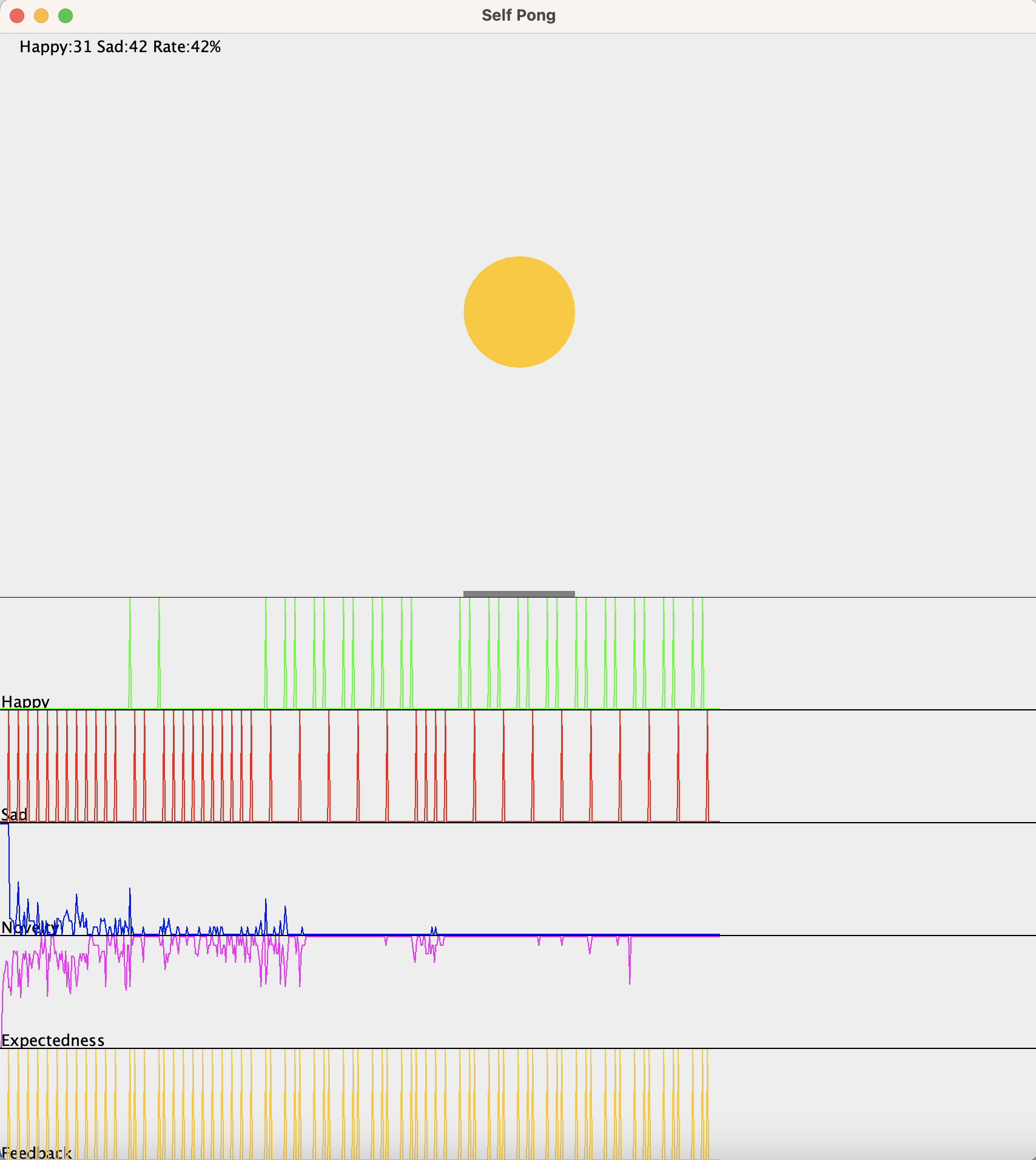}
\caption{Visualization of the process of learning to play ping-pong against the opposite wall in the space of needs: the opposite wall is on top, the racket is in the middle, the ball is between the wall and the racket, at the bottom are the plots rendering functions of satisfaction or dissatisfaction for four needs: Happiness ("Happy"), Sadness ("Sad"), "Novelty", "Expectedness"; and a separate function of explicit (positive or negative) reinforcement presence ("Feedback").}
\label{fig5_self_pong}
\end{figure}

In the experiment, the agent's behavior was carried out in a four-dimensional need space: 1) positive reinforcement for bouncing the ball or hitting it against a wall ("Happy"); 2) avoiding negative reinforcement for hitting the ball against the wall on the player with the racket side ("Sad"); 3) the novelty of the states detected by the agent ("Novelty"); 4) the expectation or predictability of experienced situations ("Expectedness"). In this case, the actualization $y$ of all four needs was measured and visualized (Fig.~\ref{fig5_self_pong}), and different values of the priority vectors $x$ for the first two of them, associated with reinforcement $r$, were also investigated. The experiment was conducted using various experience-based learning strategies, including both explicit reinforcement through several external channels and the satisfaction of the agent's internal needs, described previously \cite{kolonin2022}. In this case, the learning strategies involved forming episodic memory either as segments of $s_t$ sequences (including $a_t$ actions) between reinforcements (positive and negative) or as maps of transitions between states $s_{\tau-1}$ and $s_{\tau}$. In both cases, according to the "global feedback" ("global reinforcement") principle \cite{kolonin2022}, reinforcement $r_t$ was used to change the utility values of all episodic memory elements in temporal segments between reinforcements, in contrast to the Q-learning method \cite{li2023reinforcement}, where reinforcement changes gradually, propagating along the time axis from future states to previous ones.

The experiment confirmed the effect of negative reinforcement on the agent's learning ability, as described in \cite{kolonin2022}. That is, when the agent $x$ profile placed equal emphasis on receiving positive reinforcement and avoiding negative reinforcement, so that positive and negative feedback contributed equally to the formation of the utility $U$ within the model, learning slowed down, and for some models, game board configurations, and reinforcement delays, the acquisition of game skills became impossible due to the fact that negative feedback suppressed the agent's exploratory activity and punished its efforts spent on creating new strategies. By prioritizing receiving positive feedback over avoiding negative feedback, learning ability remained stable and was achieved across all experimental conditions.

\section{Conclusion}

We propose an anthropocentric model of the psyche with the potential to mimic human behavior or create artificial general intelligence systems for artificial agents, relying on the space of needs as the driving core of the space of states, and decision making system operating in the latter space relying on either symbolic or sub-symbolic approaches. We provide the computational concept of this model and present a memory architecture for its implementation.

To validate our concept, we present a preliminary computational implementation of a cognitive architecture based on the proposed formalism and perform a brief experimental study. This study yields preliminary results that are compelling in their interpretation and open the way for further research.

%
%
%
\bibliographystyle{splncs04}
\bibliography{psyche2026}

\section{Appendix}

\subsection{Impact Statement}

The human intellect always answers the question: "Will things be better or worse if I make this decision?" The economy satisfies human needs — physical and psychological —i n their social and productive activities. The psyche controls the satisfaction of needs at the level of individual behavior. Using an appropriate mathematical model, we can calculate the degree to which needs in a system are satisfied through the expenditure of energy and time on activity — both at the economic and mental levels. This is the primary goal of intelligence.

The foundation for solving the problem of survival is a continuous supply of vital energy. Human activity requires energy expenditure. Thus, there is an interdependence between energy acquisition and efficient use.

Intelligence enables one to choose the optimal solution. The higher the intelligence, the more effective the outcome. Increasing the collective intelligence quotient contributes to the development of civilization. The value of the intelligence coefficient indicates how much intellectual effort was previously expended to create and acquire a given object or knowledge.

To ensure the survival of a species, society, or individual, it carries out targeted activities, drawing on existing knowledge and experience, acquired individually or collectively, maximizing the intelligence coefficient of both the individual and society. In turn, when creating and comparing artificial intelligence systems, maximizing this intelligence coefficient can be considered the primary indicator of system effectiveness.

The ability of an intelligent agent to operate in complex environments, taking into account multiple goals and threats defined within the context of an individual's space of biological, physical, and existential needs, while optimizing the satisfaction of the latter, can be considered a form of general intelligence, the strength of which is determined by the dimensionality and complexity of this space.

\subsection{Possible Applications}

One of the most intuitive practical applications can be found in any process control system, subsystem, such as any device where a stream of state variable values from the field is used to generate a stream of control signals that maintain a controlled environment, such as a plant, engine, vehicle, or any piece of equipment, in the state desired by the operator or entire system, or move it to a desired state. In real-world industrial applications, this type of control currently cannot be delegated to uninterpreted solutions based on modern neural networks and is performed either manually or using hand-written software.

For example, the application of such a solution in an industrial automation environment can occur either at the lower level of industrial controllers, learning patterns based on their programmable logic operation and generate alerts if the operation deviates from these patterns due to error or malfunctioning or unexpected condition, or at the upper level of the human-machine interface, where such a solution can also generate alerts in unexpected situations or provide timely recommendations to operators.

\subsection{Future Work}

The next important steps on our road-map include improving the stability of training, implementing environment-independent dimensionality reduction and testing it in various environments, enabling the partitioning of the state space into subspaces corresponding to multiple simultaneous actions, experimenting with account for real energy-efficiency, and moving toward practical applications such as real-world control systems, such as industrial automation or smart homes, where hierarchically dependent state subspaces may exist. Another challenging task is extracting programmable logic, transparent, understandable and configurable by humans, from the resulting interpretable models from the state transition graph model using an extended version of the "super-compilation" approach.

\end{document}